\documentclass{article}

\usepackage[final, nonatbib]{neurips_2020}
\usepackage[numbers, sort]{natbib}

\usepackage[utf8]{inputenc} 
\usepackage[T1]{fontenc}    
\usepackage{hyperref}       
\usepackage{url}            
\usepackage{booktabs}       
\usepackage{amsfonts}       
\usepackage{nicefrac}       
\usepackage{microtype}      

\usepackage{graphicx}
\usepackage{amsmath}
\usepackage{svg}
\usepackage{subcaption}
\usepackage{enumitem}
\setlist{leftmargin=5.5mm}
\usepackage[font=small, labelfont=bf]{caption}
\usepackage[colorinlistoftodos]{todonotes}
\usepackage[title]{appendix}
\title{Improving Interpretability in Medical Imaging Diagnosis using Adversarial Training}

\author{%
  Andrei Margeloiu$^1$, Nikola Simidjievski$^1$, Mateja Jamnik$^1$, Adrian Weller$^{1,2}$\\
  $^1$University of Cambridge, UK\\
  $^2$The Alan Turing Institute, UK\\
  \texttt{[am2770,ns779,mj201,aw665]@cam.ac.uk}\\
}

\begin{document}
\maketitle
\begin{abstract}
We investigate the influence of adversarial training on the interpretability of convolutional neural networks (CNNs), specifically applied to diagnosing skin cancer. We show that gradient-based saliency maps of adversarially trained CNNs are significantly sharper and more visually coherent than those of standardly trained CNNs. Furthermore, we show that adversarially trained networks highlight regions with significant color variation within the lesion, a common characteristic of melanoma. We find that fine-tuning a robust network with a small learning rate further improves saliency maps' sharpness. Lastly, we provide preliminary work suggesting that robustifying the first layers to extract robust low-level features leads to visually coherent explanations.
\end{abstract}

\section{Introduction}
As diagnostic errors contribute to approximately 10\% of patients' deaths and up to 17\% of hospital adverse events \cite{national2015improving}, there is great hope for machine learning to help in medical diagnosis \cite{murdoch2013inevitable}. However, lack of transparency or inaccurate explanations for model predictions may lead to suboptimal or even harmful treatments \cite{arrieta2020explainable}.

We explore the extent to which adversarial training \cite{madry2018towards} could improve the interpretability of deep neural networks, specifically in detecting skin cancer. Adversarially trained classifiers learn robust features and can synthesize realistic images by iteratively updating the pixels in order to maximize the score of a target class \cite{ilyas2019adversarial, tsipras2018robustness, engstrom2020adversarial}. This holds even when the classifier has low robustness to adversarial attacks \citep{aggarwal2020on}. Regarding adversarially trained classifiers' interpretability, \citet{zhang2019interpreting} showed that they provide more visually coherent SmoothGrad saliency maps \citep{smilkov2017smoothgrad} and more shape-based saliency maps on natural images in general.

However, these claims arise from experiments on natural images, which may differ significantly from medical images. We suspect that distinguishing medical pathologies involves a careful inspection of small changes in textures across a particular region, to a greater extent than natural images. Moreover, there are typically far fewer images in medical datasets. Hence, we want to investigate if adversarial training can also be used on smaller size medical imaging datasets to improve interpretability.

\textbf{Contributions:} 
To the best of our knowledge, we present the first study on using adversarial training to improve interpretability in a medical imaging machine learning setting. We show that adversarially trained models have sharper and more visually coherent gradient-based saliency maps, highlighting the changes in the lesions' color, a common characteristic of melanoma. Further, we present initial findings suggesting that standard fine-tuning an already adversarially trained model further sharpens the saliency maps.

\section{Method and Experimental Setup} \textbf{Adversarial training:} Classifiers that achieve high accuracy on adversarially perturbed inputs \cite{szegedy2014intriguing} are called adversarially robust (or simply \textbf{robust}). A common way to obtain robust classifiers is using adversarial training \cite{madry2018towards}, which approximates the solution to a min-max objective function. To apply adversarial training, at every training iteration, the training samples are augmented with an adversarial perturbation. Typically the set of allowed perturbations, $\Delta$, is a norm ball $\Delta=\{\delta:\|\delta\| \leq \epsilon\}$ (using e.g., $l_1$, $l_2$, or $l_{\infty}$ norm), where $\epsilon$ is the maximum size of the norm.

\textbf{Proposed method:} We propose using adversarial training as a replacement to standard training \textit{after} all training hyper-parameters have been selected in a standard way. Thus, using adversarial training does not require changing the model architecture or hyper-parameters and can be added later if improved interpretability is desired.
\begin{enumerate}[itemsep=0.2pt,topsep=0pt]
\item \textit{Perform hyper-parameter tuning} using standard training.
\item \textit{Retrain the same model from scratch using adversarial training and re-use the same hyper-parameters previously found.} We suggest using a PGD adtversary \citep{madry2018towards} with an $l_2$ norm $\epsilon$. The adversary power $\epsilon$ should be tuned depending on the dataset, but we suggest using $\epsilon=4$ for 50,000 image pixels (scale $\epsilon$ proportionally to the desired input size). Perform perturbation augmentation using seven steps of size $\epsilon/5$ and start from a different random point. The adversarial image is clipped in the allowed range [0, 1]. Note that all other training hyper-parameters are the same as found using standard training in step 1.
\end{enumerate}

\textbf{Experimental Setup:} We perform experiments in an image classification problem, on a subset of the dermatology dataset HAM10000 \cite{tschandl2018ham10000} containing three classes: Melanoma, Melanocytic Nevi, and Benign Keratosis. We use a ResNet-18 \cite{he2016deep} architecture. All relevant code is available at \url{https://github.com/margiki/Interpretability-Adversarial}. See Appendix A for training details.

\textbf{Saliency maps} are a common way to understand the predictions of a model by highlighting the features (pixels in case of images) deemed important to the model in making the prediction. A more important feature has a higher color intensity. All showcased saliency maps are computed on correctly-classified images. Appendix B presents in detail common saliency methods.

\textbf{Evaluation:} We follow previous work in interpretability \cite{sundararajan2017axiomatic, smilkov2017smoothgrad} and \textit{qualitatively} assess the \textit{visual coherence} of saliency maps of robust and standard models. Following \citet{smilkov2017smoothgrad}, we use \emph{visual coherence} to mean that salient areas primarily highlight the object of interest, rather than the background. Evaluating models' interpretations remains an active area of research. Current quantitative metrics \cite{hooker2019benchmark, yang2019benchmarking, samek2016evaluating} seem unsuitable here since they are based on performing various input perturbations (which make less sense for adversarially trained models).

\begin{figure}[t!]
    \centering
    \includegraphics[width=0.65\textwidth]{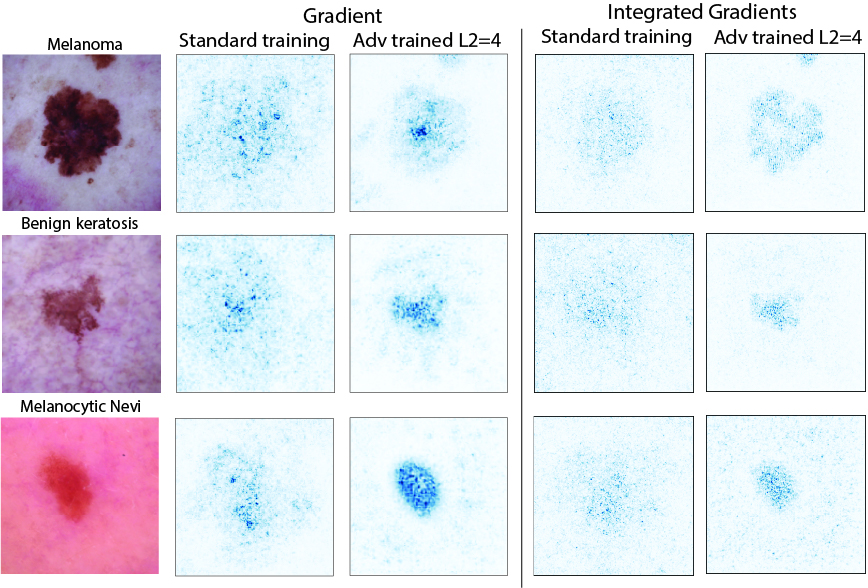}
    \caption{Saliency maps (obtained using Gradient and Integrated Gradients (IG) with Uniform baseline) on a standard and a robust model. Notice that the robust model has significantly sharper saliency maps, and it highlights the skin lesion predominantly. Furthermore, notice that the saliency map using IG of the robust model (last column of the first row) highlights the color changes in the lesion, common to melanoma.}
    \label{fig:gradient_ig}
\end{figure}

\section{Results and Discussion}
\noindent
\textbf{Robust models have sharper and more visually coherent saliency maps.} Figure \ref{fig:gradient_ig} shows that robust models provide sharper and more visually coherent gradient-based saliency maps using Gradient \cite{sundararajan2017axiomatic} and downstream methods such as Integrated Gradient (IG) \cite{sundararajan2017axiomatic}. Notice that the high saliency values are attributed predominantly inside the skin lesion, which is the object of interest. See Appendix B for more figures, including positive results using perturbation-based attribution methods.

\noindent
\textbf{Robust models may learn robust features from medical images.} In the results on melanoma using Integrated Gradients from Figure \ref{fig:gradient_ig} (last two columns of the first row), notice that the saliency map of the standard model does not resemble the change in variation from the lesion. However, the robust model (last column of the first row) predominantly highlights regions with significant color variation, a common characteristic of melanoma. This indicates that robust models may indeed learn to make predictions based on robust features \cite{ilyas2019adversarial}.

\begin{figure}[h!]
    \centering
    \includegraphics[width=1\textwidth]{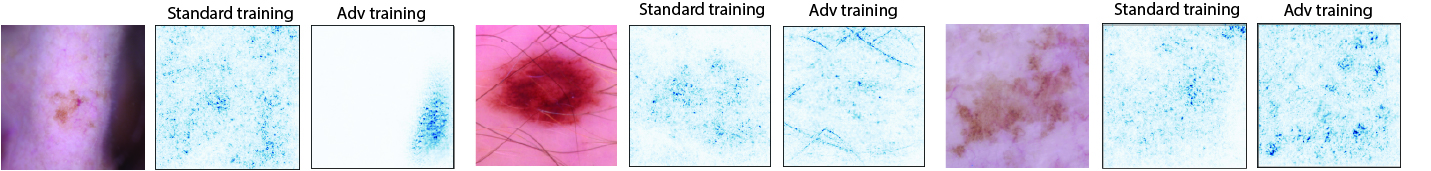}
    \caption{Fail-cases of the saliency maps of robust models. Notice that artifacts such as dark regions (lesion 1) or hairs (lesion 2) can be highlighted as relevant, or the saliency map can be noisy (lesion 3).}
    \vspace{-5pt}
    \label{fig:failcases}
\end{figure}

\noindent
\textbf{Limitations:} Figure \ref{fig:failcases} shows that the saliency maps of robust models can highlight image artifacts such as dark regions or hairs, as well as being noisy. On further investigation, we found that the maximum perturbation size $\epsilon$ used for training greatly influences the final saliency map. This can be caused partially because saliency methods are not stable to training noise \cite{smilkov2017smoothgrad} (empirical examples available in Appendix C.1 and C.2). However, our early results indicate the possibility of an image-specific optimal training perturbation $\epsilon$ for providing the sharpest saliency maps (more details about this hypothesis in Appendix C.3).

It will be interesting to explore how well our claims generalize to 
other types of medical images (e.g., X-rays, MRI scans), datasets, and network architectures.

\begin{figure}[h!]
    \centering
    \includegraphics[width=1\textwidth]{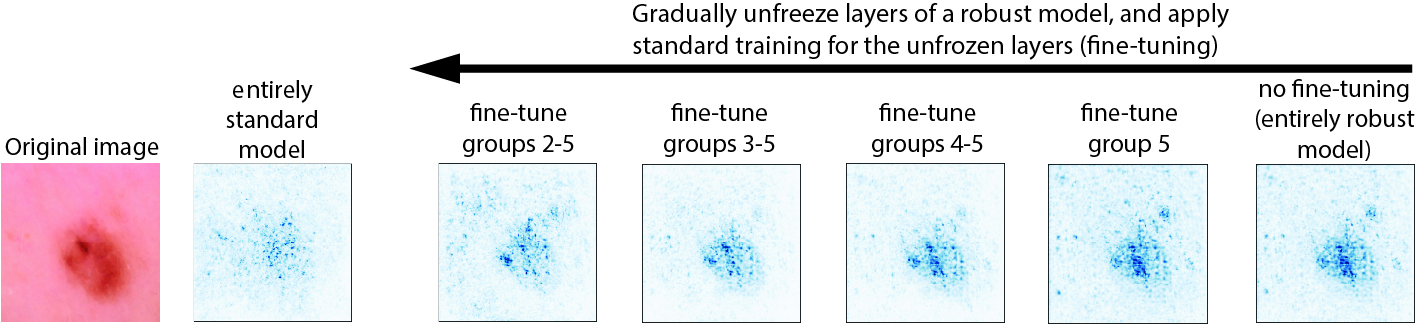}
    \caption{Gradient saliency maps when gradually unfreezing the layers of a robust network, and performing standard fine-tuning. The last column shows the saliency maps of a robust model. Columns 3-7 represent new models obtained by fine-tuning a robust model (from the last column) with a small learning rate. By fine-tuning the last layers, we essentially enable them to combine the robust features extracted by the first layers. Notice how fine-tuning the last layers reduces the noise in the saliency maps.}
     \vspace{-5pt}
    \label{fig:fine-tuning-main}
\end{figure}

\textbf{Conclusion and Future Work:} We presented a preliminary investigation on a medical imaging dataset, suggesting that gradient-based saliency maps can become sharper and more visually coherent using adversarial training. This way, robust models attribute high saliency values predominantly to the lesion and its color change, common to melanoma.

This work suggests several directions for future research. We intend to investigate which layers are important to `robustify' to obtain visually coherent explanations. Figure \ref{fig:fine-tuning-main} shows early evidence that fine-tuning an already robust network further sharpens the saliency maps - suggesting that having robust first layers to extract robust low-level features leads to visually coherent explanations (see training details and more figures in Appendix D). Finally, we look forward to seeing how our findings generalize to other datasets and network architectures.

\section*{Broader Impact}
Convolutional Neural Networks are applied in a number of classification tasks involving Medical Images. As diagnostic errors contribute to approximately 10\% of patients' deaths and up to 17\% of hospital adverse events \cite{national2015improving}, having a reliable diagnosis is critical. Our research can help mitigate misdiagnosis by enabling doctors to better individual decisions of the model and providing trustworthy explanations.

The potential risks of our method are associated with the doctors becoming overconfident in their decision, thus not seek a second opinion and lead to misdiagnosis. To address this issue, we clearly mentioned the limitations of our work and made specific that it can capture artifacts (see Figure \ref{fig:failcases} and common fail-cases in Appendix C). The proposed method is not ready to be used in a real-world scenario. We encourage research in understanding why saliency methods are brittle to training noise as well as understanding why saliency method can capture artifacts (e.g., dark regions in the skin lesion). Furthermore, the used dataset contains images of only one skin type; thus, investigating more diverse datasets is required to generalize the method's effectiveness on other skin types.

As a near-term impact, we envision researchers and practitioners using this method for "bug-fixing" during model experimentation to understand fail-cases better and improve their models.

To advance the collective understanding of the limitations of using adversarial training to improve interpretability, we encourage research in understanding the explanations of robust models in real-world scenarios in Medicine, other high-stake domains (e.g., autonomous driving). We also acknowledge the importance of gathering more diverse datasets so that progress in AI benefits all people.

\begin{ack}
Andrei Margeloiu acknowledges support from the Cambridge ESRC Doctoral Training Partnership. Nikola Simidjievski and Mateja Jamnik acknowledge support from The Mark Foundation for Cancer Research and Cancer Research UK Cambridge Centre [C9685/A25177]. Adrian Weller acknowledges support from the David MacKay Newton research fellowship at Darwin College, The Alan Turing Institute under EPSRC grant EP/N510129/1 and U/B/000074, and the Leverhulme Trust via CFI. 


\end{ack}

\bibliographystyle{plainnat}
\bibliography{references}

\clearpage
\begin{appendices}
\section{Experimental Setup and Training details}

\begin{figure}[h!]
\begin{minipage}[b]{\textwidth}
  \begin{minipage}[l]{0.62\textwidth}
  		\centering
  		\includegraphics[width=\textwidth]{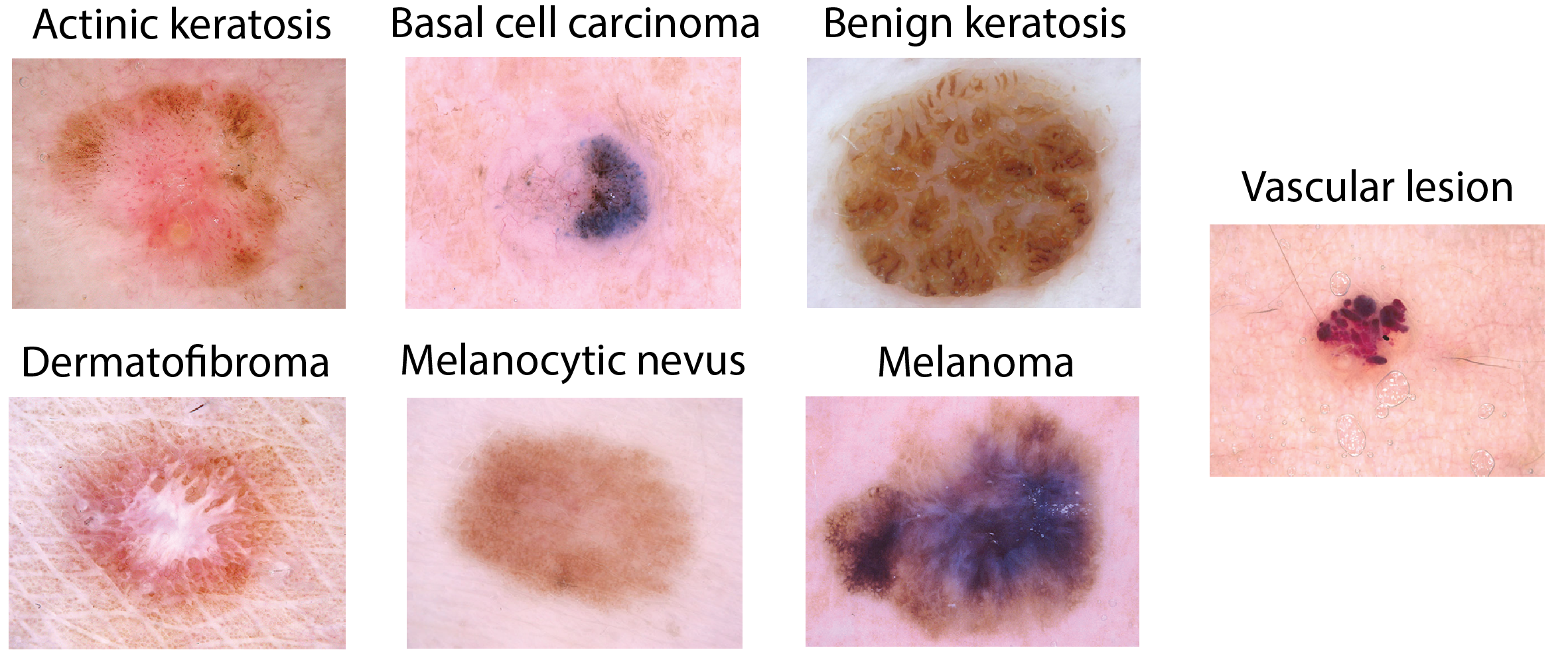}
        \captionof{figure}{Overview of the dataset. Left: Images from each class of the dataset. Right: Dataset distribution.}
        \label{fig:ham10000}
  \end{minipage}%
  \hfill
  \begin{minipage}[r]{0.34\textwidth}
    	\centering
		\resizebox{0.8\columnwidth}{!}{%
        	\begin{tabular}{@{}ll@{}}%
				\toprule
				Lesion type           & Images \\ \midrule
				Melanocytic nevi      & 6705   \\
				Melanoma              & 1113   \\
				Benign keratosis & 1099   \\
				Basal cell carcinoma  & 514    \\
				Actinic keratosis     & 327    \\
				Vascular lesion      & 142    \\
				Dermatofibroma        & 115    \\ \bottomrule
			\end{tabular}%
		}
	  \label{table:ham10000}
    \end{minipage}
  \end{minipage}
\end{figure}

\textbf{Dataset:} We use the dermatology dataset HAM10000 \cite{tschandl2018ham10000} which contains 10,015 images presenting seven lesion types as shown in Figure \ref{fig:ham10000} with the number of instances of each. Adversarial training studies are not typically conducted on datasets with high-class imbalance (the majority/minority ratio is $58:1$ for this dataset). Consequently, we perform experiments on a randomly selected balanced subset of three classes: Melanocytic Nevi, Benign Keratosis, and Melanoma. We randomly split the dataset into training (2400 images: 800 for each class) and test (897 images: 299 for each class). The training set is randomly split into five folds, which are used for cross-validation. There is no overlap between training and test sets or between the folds in cross-validation.

\textbf{Architecture:} We use a ResNet-18 architecture \cite{he2016deep}, with the weights initialised from a model pretrained on ImageNet \cite{imagenet}. The last fully connected layer is replaced with one that has three outputs (i.e., one for each target class), and whose weights are initialized using Kaiming initialization \cite{he2015delving}. 

\textbf{Training:} We use the transfer learning methodology first to train the randomly initialized last layer, and then we fine-tune the entire network. All training is performed using Adam \cite{kingma2015adam} with standard parameters ($\beta_1 = 0.9$, $\beta_2 = 0.999$) and with weight decay of 5e-4. We train only the last fully connected layer for ten epochs (to avoid large gradient updates that would collapse the pre-trained weights in the first convolutional layers during fine-tuning), with a mini-batch of size 32 and learning rate $\alpha = 0.001$. The fine-tuning is performed as follows.

We fine-tune the whole network for 25 epochs, with a mini-batch of size 16 and a learning $\alpha = 0.0003$. The learning rate is decayed by a factor of 10 after 15 epochs. The fine-tuning hyperparameters were found using five fold cross-validation on the training set, through a grid-search of learning rates of $1e-3, 3e-4, 1e-4$ and mini-batch sizes of 8, 16, 32. We selected the hyper-parameters providing the highest accuracy and trained the final model on the entire training set.

For the fine-tuning experiment (Appendix D), we randomly take 150 samples as a balanced validation dataset for early stopping, leaving 747 samples for the test dataset. The hyper-parameters and training procedure are as presented above and have two changes to accommodate for extra fine-tuning a robust network. Firstly, we use early stopping if the validation loss does not decrease for ten epochs. Secondly, we decrease the learning rate by a factor of 10 if the validation loss does not decline for five epochs.

\textbf{Adversarial training:} In all experiments, we obtain robust classifiers by closely following the adversarial training methodology of \citet{madry2018towards} by augmenting the training data with perturbations coming from a projected gradient descent (PGD) adversary. Thus, we train against a PGD adversary in the $l_2$ norm using seven steps, varying power $\epsilon$ (mentioned in each experiment), step size of $\epsilon/5$ and starting from a different random point. The adversarial image is clipped in the allowed range [0, 1]. All other hyper-parameters are the same as during standard training.

\textbf{Data augmentation:} The original images have size 600x450px. We crop the centre region of 450x450px and downscale to 224x224px. We apply data augmentation in order to prevent overfitting: random horizontal/vertical flips, changing the brightness, contrast, and saturation randomly by up to $0.2$, and applying random affine transformation of $10^{\circ}$ and rotation of up to $50^{\circ}$. The images are normalized with the mean and standard deviation of the ImageNet training set.

\section{Saliency Methods and Additional Results}
All showcased saliency maps are computed on correctly-classified images and are visualized as single-colored heatmaps. The saliency of one pixel is the sum of the absolute values of each RGB channel. We cap the extreme value to the $99^{th}$ percentile as proposed by \citet{smilkov2017smoothgrad}, and rescale the values in [0, 1].

\begin{figure}[h!]
    \centering
    \includegraphics[width=\textwidth]{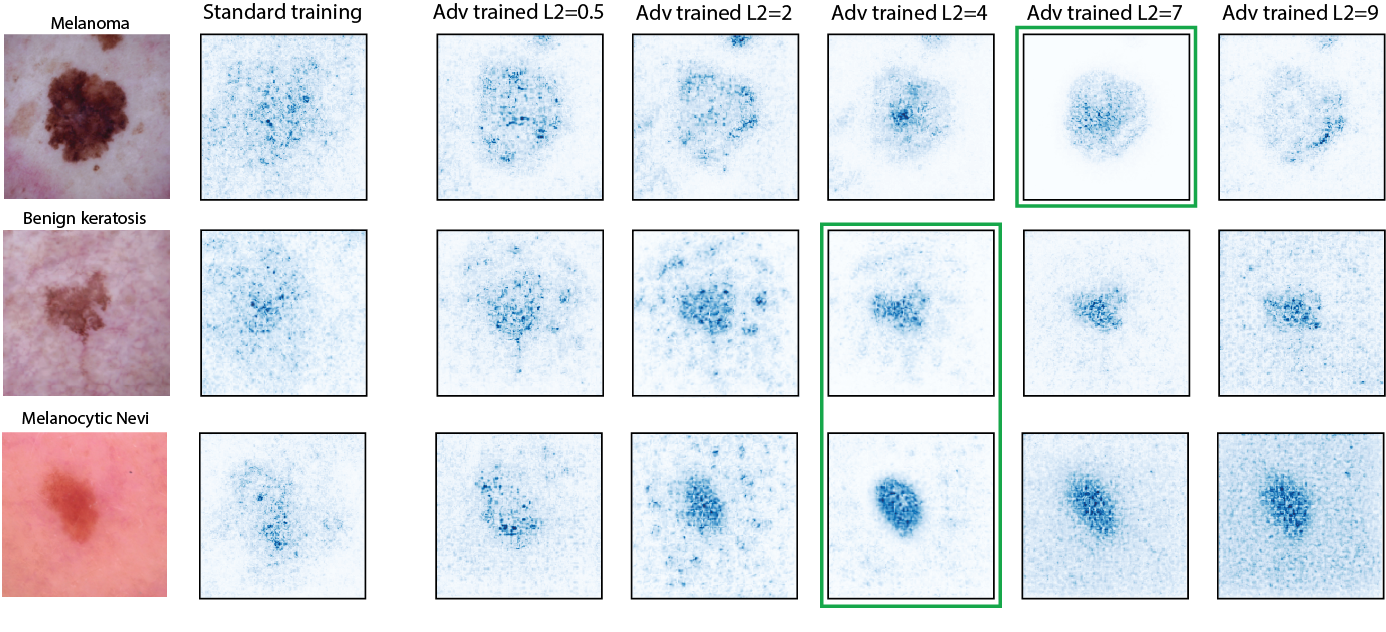}
    \caption{Gradient (also called Saliency) attribution method applied to several models. The second column shows the saliency map of a standard model. The following columns show the saliency map of robust models trained with a PGD adversary with varying $l_2$ norm. Notice that powerful adversaries ($\epsilon=4$, $\epsilon=7$) correlate with more perceptually-aligned gradients on the skin lesions. In green are highlighted the most coherent saliency maps to our opinion. Furthermore, notice that using a large adversary ($\epsilon=9$) starts showing noise in the saliency map, but still retains the perceptually-aligned gradients.}
    \label{fig:saliency_tablou}
\end{figure}

\subsection{Gradient}
Gradient (also called Saliency) \cite{simonyan2013deep} is a feature attribution method that assigns importance by computing the gradient of the score of a class of interest  $y^c$ with respect to every input feature $x_i$. For images, a feature $x_i$ represents one RGB channel of a pixel.
\begin{align}
\label{eq:saliency}
\frac{\delta y^{c}}{\delta x_i}
\end{align}

\noindent
The saliency map is constructed by highlighting each pixel proportionally to the absolute value of its gradient (eq. \ref{eq:saliency}). It is worth emphasizing that the saliency map is created for a target class $c$.

\vskip 0.4cm
\noindent
Figure \ref{fig:saliency_tablou} shows a visual comparison of Gradient applied on standard and robust models trained with varying $l_2$ norm. Notice in the second column that the saliency of a standard model is visually noisy and not very coherent with the lesion. In comparison, increasing the adversary level ($\epsilon=0.5$, $\epsilon=2$, $\epsilon=4$) makes the saliency map more coherent by having perceptually-aligned gradients with the lesion. A very powerful adversary with $\epsilon=9$ starts introducing noise.

\subsection{Integrated Gradients}
Integrated Gradients \cite{sundararajan2017axiomatic} compute the importance score by summing the gradients on images interpolated between a baseline $x'$ and the input image $x$. The baseline image $x'$ represents the absence of the input features, and we will shortly discuss several options about choosing the baseline. Thus, by integrating over the path between the baseline $x'$ and the real input $x$, we obtain multiple estimates of the importance of every feature; this avoids the issue of local gradients being saturated. Formally, the importance score for feature $i$ is computed as:
\begin{align}
    \phi_{i}^{I G}\left(f, x, x^{\prime}\right) = \left(x_{i}-x_{i}^{\prime}\right) \times \int_{\alpha=0}^{1} \frac{\delta f\left(x^{\prime}+\alpha\left(x-x^{\prime}\right)\right)}{\delta x_{i}} d \alpha
    \label{eq:ig}
\end{align}

\noindent
where $f$ represents the model, $x_i$ represents one feature of the input, $\delta$ represents partial derivative, and $\alpha$ is part of the integral and defines the distance on the path between $x$ and $x'$.

\vskip 0.4cm
\noindent
The \textbf{baseline} aims to describe the absence of the input features. Choosing the right baseline is closely related to finding a natural way of expressing the absence of the features from a dataset. With regard to images, pixels cannot be removed, and we need to define a value that represents their absence. For example, in diabetic retinopathy, a common baseline is a black image \cite{sayres2019using}.

\vskip 0.4cm
\noindent
For our skin cancer dataset, a \textbf{black baseline} is not appropriate because the black color is strongly associated with melanoma. Thus the sensitivity of the lesion's pixels would be minimal because the baseline is very similar to the lesion color. With regards to the equation \ref{eq:ig}, a black baseline makes the difference between baseline and input $(x - x') \rightarrow 0$, thus attributing zero importance to the lesion.

\begin{figure}[t!]
    \centering
    \includegraphics[width=\textwidth]{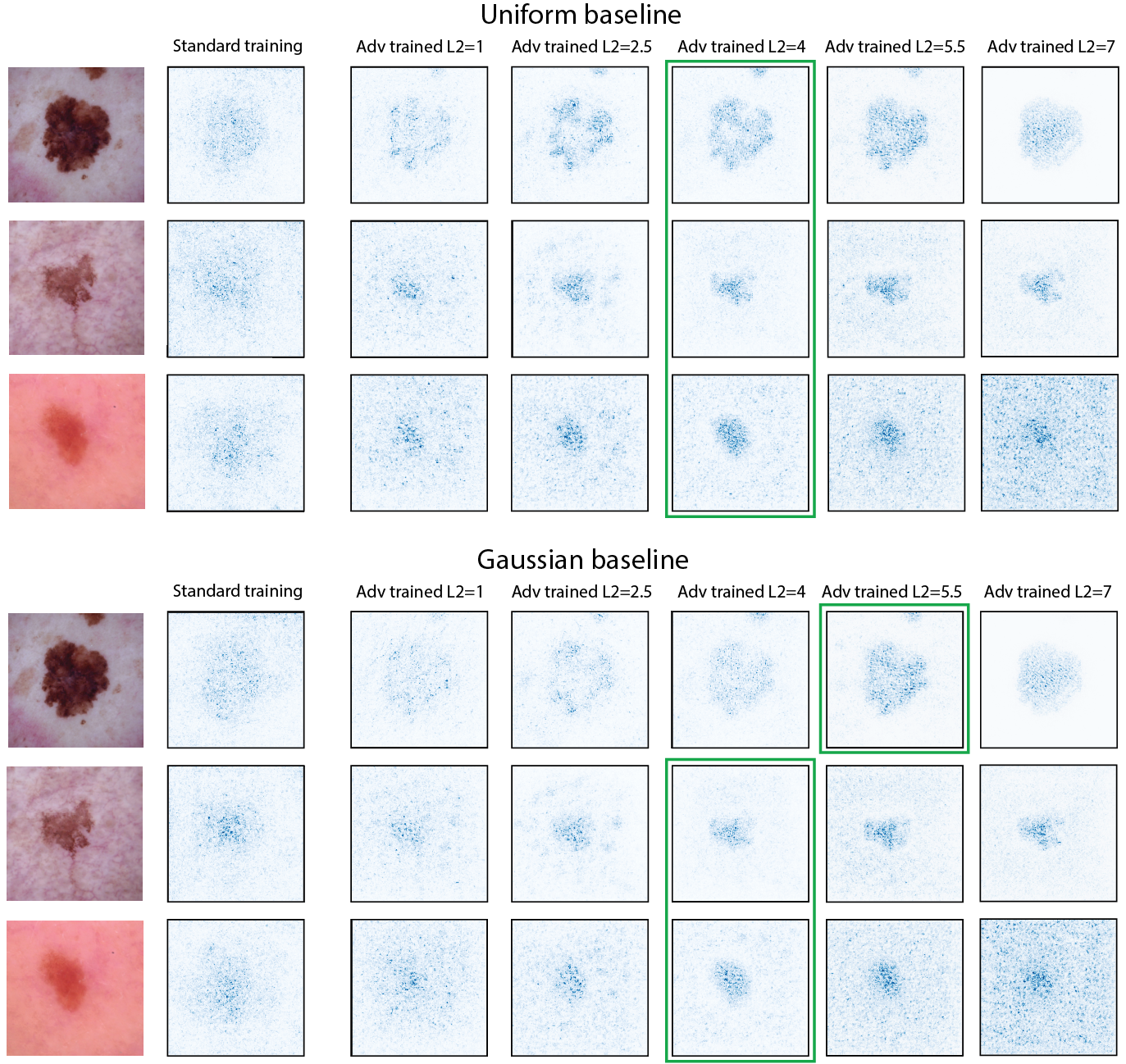}
    \caption{Integrated Gradients attribution applied on several images. The top section shows the attributions using a baseline of uniform noise, and the bottom section shows the attribution using a baseline of a Gaussian distribution centered at the original image with $\sigma=1$. The most coherent saliency maps are highlighted in green. Notice that for both baselines, robust models have more coherent saliency maps.}    	      
    \label{fig:ig_appendix}
\end{figure}

\vskip 0.4cm
\noindent
Two alternative baselines are uniform random and Gaussian \cite{sturmfels2020visualizing}:

\begin{itemize}
\item The \textbf{uniform baseline} represents an image where each pixel is drawn from a uniform distribution between [0, 1]. Intuitively, this signifies a random image that looks like noise.
\item The \textbf{Gaussian baseline} is inspired by \cite{smilkov2017smoothgrad} and is a sample from a Gaussian distribution centered at the original image with a standard deviation of $\sigma=1$. Intuitively, it can be thought of adding Gaussian noise to the input image.
\end{itemize}

\vskip 0.4cm
\noindent
Figure \ref{fig:ig_appendix} shows the attribution with a uniform and a Gaussian baseline. Notice that robust models assign importance predominantly within the boundary of the lesion; this hints that robust models make the prediction mainly by looking at the lesion.

\subsection{Occlusion}
\begin{figure}[h!]
    \centering
    \includegraphics[width=\textwidth]{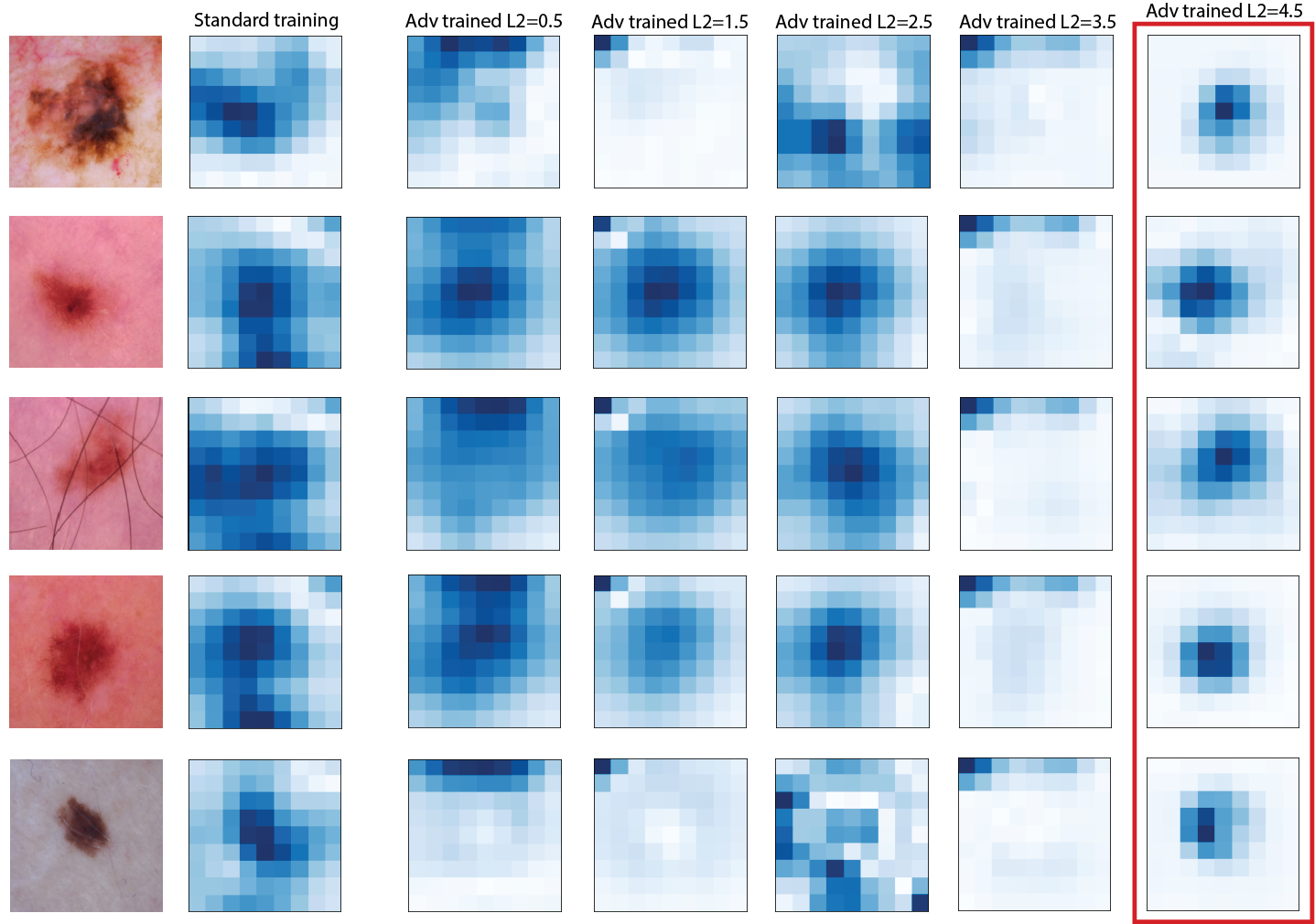}
    \caption{Occlusion attribution on a standard and several robust models. Notice that the standard model attributions are uninformative. Furthermore, it is common that robust models explanation are not coherent (columns 3-6). However, notice in the last column that the robust model trained against an adversary of $\epsilon = 4.5$ gives good localization on all images, hinting that there may be an optimal adversary.}
    \label{fig:occlusion}
\end{figure}

\noindent
Occlusion \cite{zeiler2014visualizing} is a perturbation-based approach for computing the attributions. It involves replacing rectangular patches of the input with a baseline and computing the decrease of the predicted probability for the new input. A larger drop indicates that the replaced region contained features important for the prediction. In our experiments, we use patches of 50x50px with a stride of 25px. If a feature is part of multiple patches, then it is assigned the average of each patches' sensitivity. 

\vskip 0.4cm
\noindent
Figure \ref{fig:occlusion} shows Occlusion applied to several robust models and a standard model. Firstly looking at the standard model, notice that the saliency map is noninformative because it gives non-trivial sensitivity to marginal regions of the image.

\vskip 0.4cm
\noindent
Analyzing the robust model, notice that using adversary $\epsilon=4.5$ gives good localization across several images. This suggests that the robust model learns to predict solely based on the lesion; however, we cannot extrapolate this to all robust models, since for example, for $\epsilon=0.5$, $\epsilon=1.5$, $\epsilon=2.5$, $\epsilon=3.5$ the saliency maps are not coherent.

\subsection{Grad-CAM}

\noindent
Grad-CAM \cite{selvaraju2017grad} is a technique for highlighting the regions which positively impact the prediction of a target class. Given a user-chosen convolutional layer $A$, it computes the gradient of the target class $y^c$ with respect to every unit in the volume outputed by the convolutional layer $A_{ij}^{k}$ ($k$ represents the feature map, and $i$, $j$ represent the position in the feature map). Then, it averages the gradients to create a coefficient $\alpha^{c}_{k}$ for the contribution of feature map $k$ to class $c$:
\begin{align}
    \alpha^{c}_{k} = \frac{1}{Z} \sum_{i} \sum_{j} \frac{\delta y^{c}}{\delta A^k_{ij}}
\end{align}

\noindent
The importance score attributed by Grad-CAM is the weighted average of the coefficients of the convolutional layer:
\begin{align}
    \text{Grad-CAM}^{c} = ReLU(\sum_{k} \alpha^{c}_{k}A^{k})    
\end{align}

\noindent
It is important to note two nuances of Grad-CAM. Firstly, the sensitivity is not assigned for an input feature (as it is the case with Gradient and Integrated Gradients). Instead, the sensitivity is attributed to a unit within a convolutional layer, which extracts a particular feature from a given region of the image. Secondly, it shows \textit{only positive} attributions because it applies ReLU on the attributions and all regions with a negative influence receive zero importance.

\begin{figure}[t!]
    \centering
    \includegraphics[width=\textwidth]{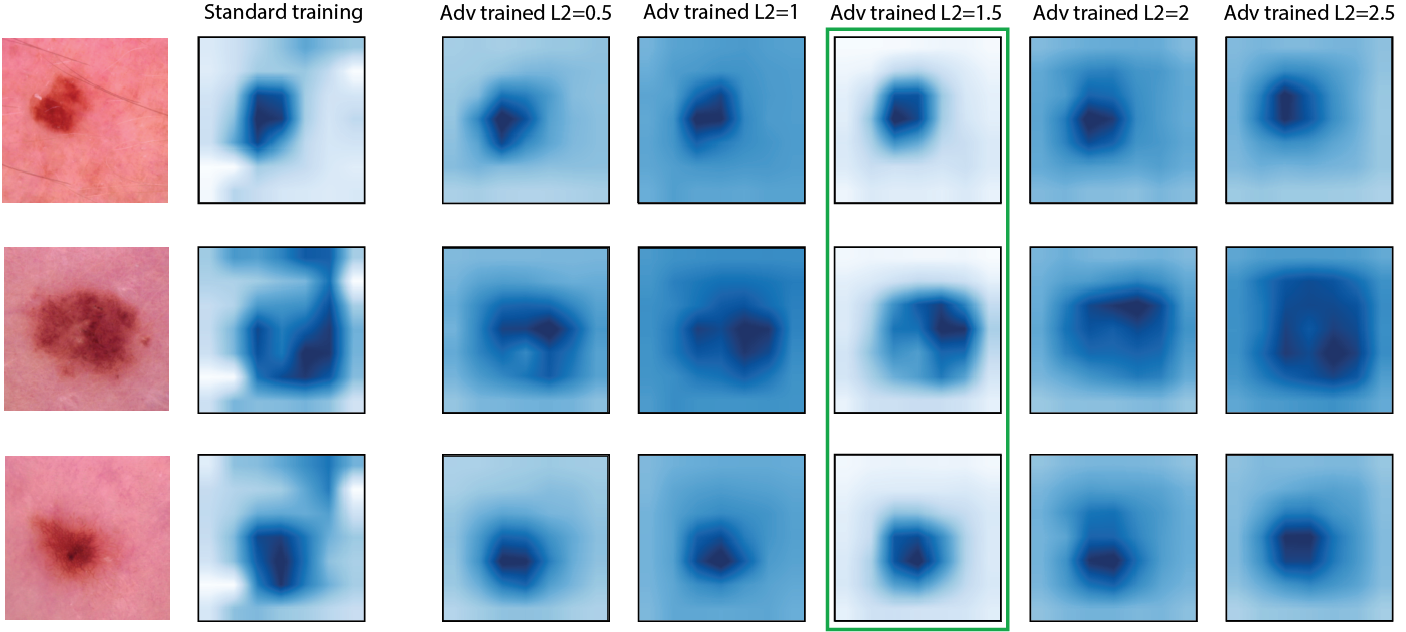}
    \caption{Grad-CAM attribution of several models; Grad-CAM highlights only the regions that positively support predicting the class of interest. Notice that the robust model with $\epsilon=1.5$ has a more informative saliency map because it gives positive attribution \textit{only} to the lesion. Other robust models attribute a non-trivial level of importance across the skin, making their explanations generally uninformative. This hints towards the possibility of an optimal value of the adversary.}
    \label{fig:gradcam_tablou}
\end{figure}

\vskip 0.4cm
\noindent
Our experiment applies Grad-CAM on the last convolutional layer of a ResNet-18 network, which has a 7x7x512 output volume. Thus the output of a Grad-CAM attribution is 7x7, where each value corresponds to a region of 224/7 = 32 pixels. Figure \ref{fig:gradcam_tablou} shows the attribution on several models. Notice that the robust model $\epsilon=1.5$ has a coherent saliency map, as it attributes importance \textit{only} to the lesion. Interestingly, the other robust models assign a non-trivial positive influence on the skin regions; however, the marginal regions should not carry much importance for the prediction. 

\clearpage
\section{Limitations}
\subsection{Additional fail-cases for saliency maps of robust models}

\begin{figure*}[h!]
    \centering
    \includegraphics[width=0.7\textwidth]{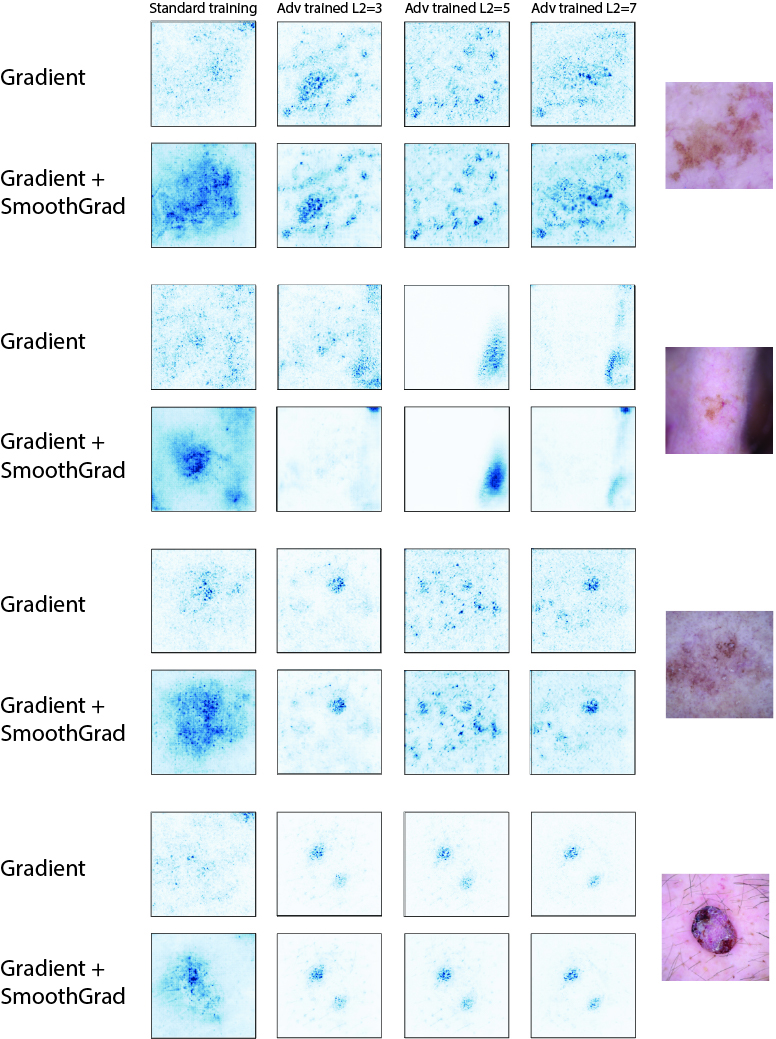}
    \caption{Inputs on which the saliency maps the robust model fail to provide meaningful explanations (e.g., highlights artifacts such as dark regions or providing noisy explanations).}
    \label{fig:grad_smoothgrad_failcases}
\end{figure*}

See Figure \ref{fig:grad_smoothgrad_failcases}.

\subsection{Saliency maps are brittle to training noise}

\begin{figure}[h!]
\centering
\begin{subfigure}{.45\textwidth}
  \centering
  \includegraphics[width=0.9\linewidth]{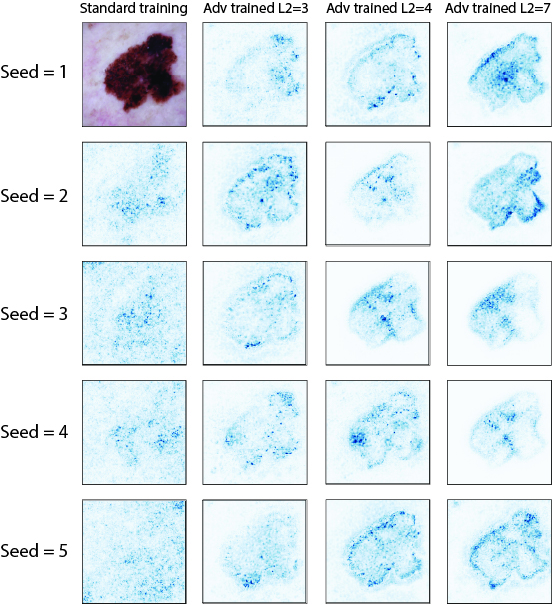}
  \caption{Saliency maps using Gradient.}
\end{subfigure}%
\begin{subfigure}{.45\textwidth}
  \centering
  \includegraphics[width=0.9\linewidth]{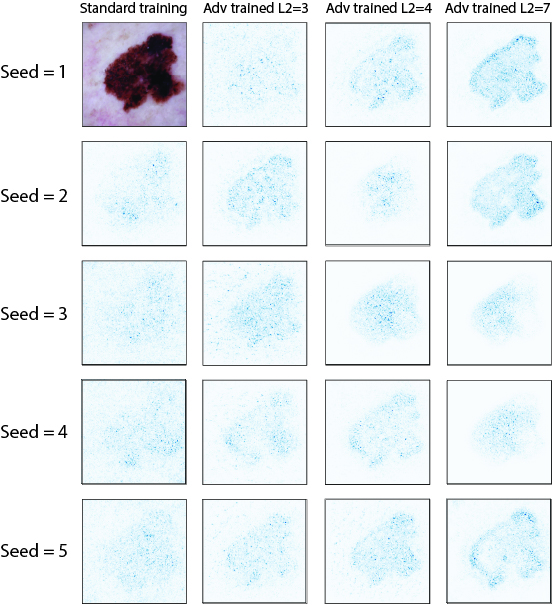}
  \caption{Saliency maps using Integrated Gradient with a Gaussian Baseline.}
\end{subfigure}
\caption{Saliency maps using Gradient and Integrated Gradient on multiple models trained identically except the random seed. The random seed influences the order of the mini-batches. Notice that the saliency maps differ in non-trivial ways across the runs, showing the saliency method themselves are brittle to training noise.}
\label{fig:brittleness}
\end{figure}

See Figure \ref{fig:brittleness}.

\subsection{Varying the adversary power}

\begin{figure*}[t!]
    \centering
    \includegraphics[width=\textwidth]{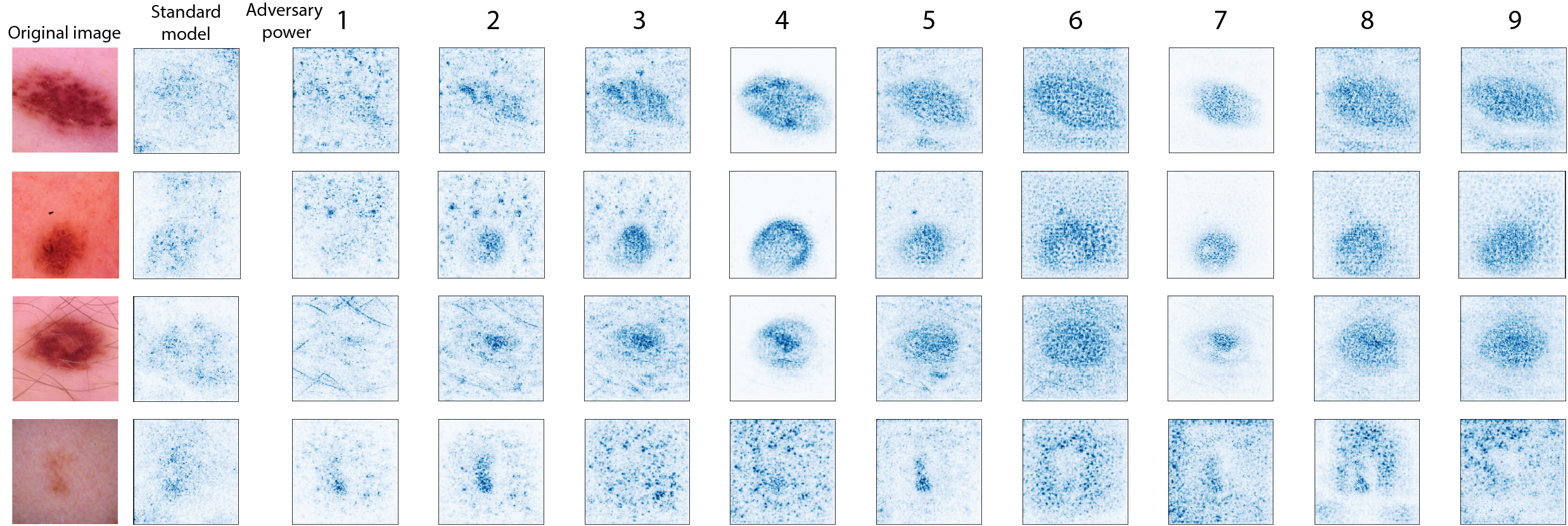}
    \caption{Each row shows the Gradient saliency maps of a standard and robust models trained with varying adversary powers in the $l_2$ norm. Notice how adversarially trained models with different adversary powers provide different saliency maps on the same input. Also, see that the robust model trained with $\epsilon=4$ gives a relatively clear explanation for the first three images, but a noisy explanation for the fourth image. This hints towards the possibility of an image-specific optimal adversary power for providing the clearest saliency maps.}
    \label{fig:optimal_adversary}
\end{figure*}

The saliency map's sharpness varies significantly with the size of the norm used during adversarial training. Figure \ref{fig:optimal_adversary} shows that robust models trained with varying adversary powers $\epsilon$ give very different saliency maps on the same input.

Our results indicate the possibility of an image-specific optimal adversary power, similar to a sweet spot, for providing the sharpest saliency maps. Thus, we propose the hypothesis: "The power of the adversary is an image-specific hyper-parameter of the explanations, where the optimal value gives the clearest explanation." It is important to note that different images have different optimal values (in Figure \ref{fig:optimal_adversary}, $\epsilon=4$ seems to be optimal only for the first three images). As a rough empirical estimation, we observed that the optimal values tend to be between $\epsilon=3$ and $\epsilon=5$, while for $\epsilon \geq 8$, the saliency maps become very noisy. 

An interesting observation emerges by looking at the third row from Figure \ref{fig:optimal_adversary}. Notice that for $\epsilon=1$, the saliency map highlights predominantly the hairs, which are irrelevant for the prediction. However, as $\epsilon$ increases, the model shifts its attention to the skin lesion, which is of interest. We leave the investigation on how to select the optimal adversary power for future work.

\clearpage
\section{Standard fine-tuning the last layers of a robust model}

\begin{figure*}[h!]
    \centering
    \includegraphics[width=\textwidth]{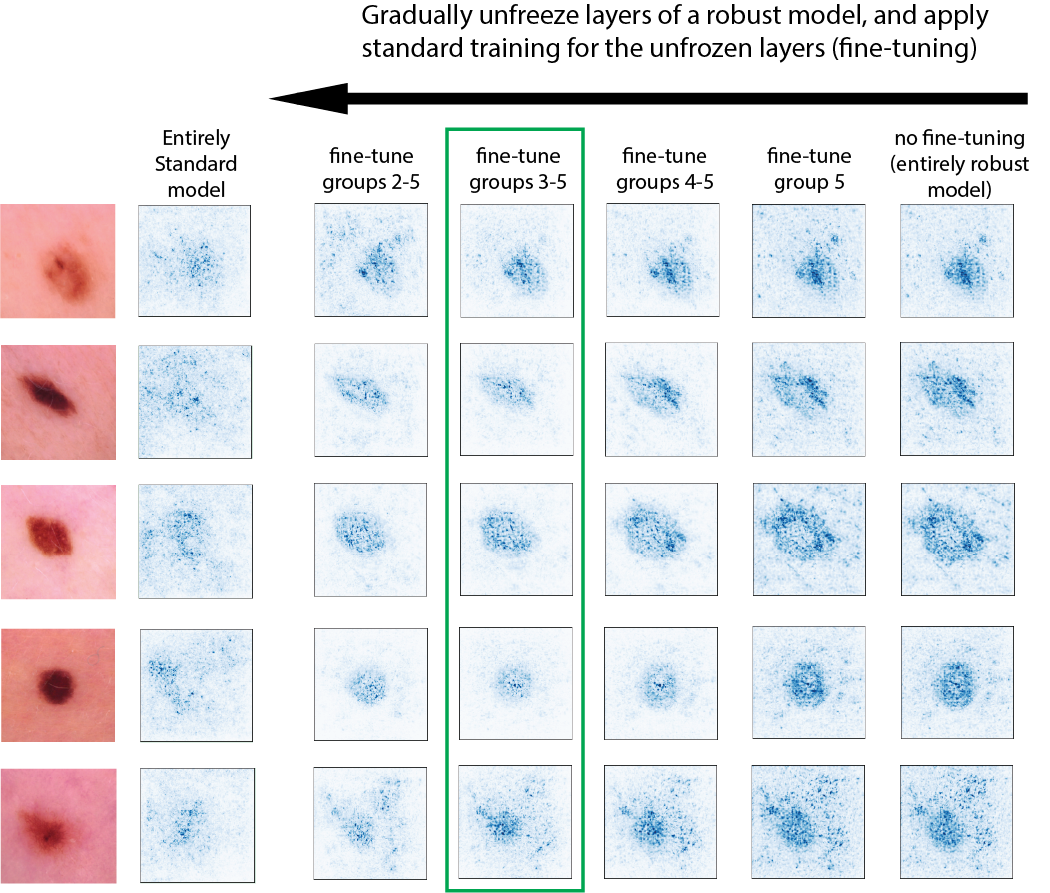}
    \caption{Saliency maps using Gradient method when gradually unfreezing the layers of a robust network, and performing standard fine-tuning. The last column shows the saliency maps of a robust model with $\epsilon = 3$. Columns 3-7 represent new models obtained by fine-tuning a robust model (from the last column) with a learning rate of $1e-5$. The layers which are not fine-tuned continue to extract robust features. By fine-tuning the last layers, we essentially enable them to combine the robust features extracted by the first layers. Notice how fine-tuning the last layers reduces the noise in the saliency maps.}
    \label{fig:fine-tuning}
\end{figure*}

\label{subsec:finetuning}
Making accurate predictions is crucial in high-stake domains, such as Medicine. However, robust models have lower predictive performance than standard models \cite{tsipras2018robustness}. This raises the question: "Can the performance of a robust model be increased while maintaining its visually coherent explanations?". The answer is affirmative, and we shortly present a methodology for achieving this.

We hypothesize that higher predictive performance could be achieved by making only some layers learn robust features, while other layers learn standard features. This could combine both sides' benefits: higher predictive performance from standard training and sharper saliency maps from adversarial training.

\textbf{Groups of layers:} The ResNet-18 architecture \cite{he2016deep} contains five groups of layers: four blocks of convolutional layers and a final fully-connected layer. We perform standard fine-tuning on some groups of layers of a robust model as follows:
\begin{enumerate}
\item Get a robust model (i.e., adversarially trained on the target dataset) and freeze all layers, such that the weights are not updated during training.
\item Unfreeze some groups of layers starting from the last one and keep the others frozen. This way, only the unfrozen layers are updated during training.
\item Perform fine-tuning using standard training with early stopping and a small learning rate (we use $1e-5$).
\end{enumerate}

\begin{figure}[h!]
    \centering
    \includegraphics[width=0.7\textwidth]{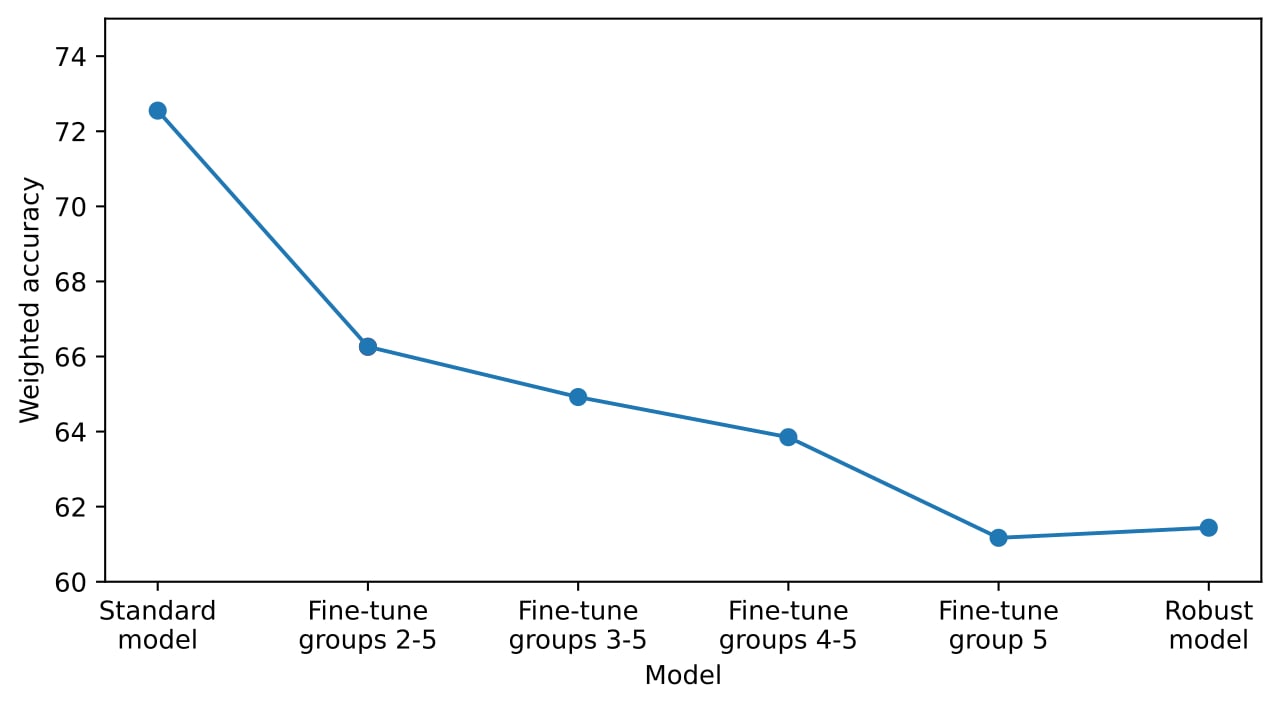}
    \caption{The accuracies of standard, robust and fine-tuned models. Notice that the accuracy increases roughly linearly with fine-tuning more layers. For example, standard fine-tuning of groups 3-5 improves the accuracy by 3\% compared to the robust model, but is still lower than the standard model.}
    \label{fig:acc_unfrozen}
\end{figure}

Interestingly, fine-tuning the last half of the network increases the predictive performance (Figure \ref{fig:acc_unfrozen}), while also reducing the noise in the saliency maps. Figure \ref{fig:fine-tuning} shows the saliency map using Gradient after unfreezing a number of groups and performing standard fine-tuning. The second column shows the standard model, and the last column shows the robust model. The middle columns show the saliency maps for models with only some of the first layers robust and the latter layers fine-tuned in a standard way.

Notice that the model which underwent fine-tuning on groups 3-5 provides a more visually coherent explanation that the standard model and a less noisy explanation than the robust model. Furthermore, it has 3\% higher accuracy than the fully robust model (Figure \ref{fig:acc_unfrozen}).

This experiment shows the potential of using robust first layers as a way to improve interpretability. A natural question is: "Why does fine-tuning reduce noise in the saliency map?". We conjecture that visually coherent explanations are attributed to having robust first layers. Each layer of a CNN can be thought of as a feature extractor at different abstraction levels by combining the features extracted by the previous layers \cite{zeiler2014visualizing}. In this way, the first layers extract robust low-level features, and the latter ones combine them in any way that increases the predictive performance. So, the fine-tuned layers learn to combine the robust low-level features extracted by the first layers. We leave this investigation as future work.
\end{appendices}

\end{document}